\documentclass[10pt,twocolumn,letterpaper]{article}

\usepackage{cvpr}
\usepackage{times}
\usepackage{epsfig}
\usepackage{graphicx}
\usepackage{amsmath}
\usepackage{amssymb}
\usepackage{multirow}
\usepackage{bm}
\usepackage{epstopdf}
\usepackage[breaklinks=true,bookmarks=false]{hyperref}

\cvprfinalcopy % *** Uncomment this line for the final submission

\def\cvprPaperID{****} % *** Enter the CVPR Paper ID here

\begin{document}

%%%%%%%%% TITLE

\title{Cross-domain Face Presentation Attack Detection via Multi-domain \\ Disentangled Representation Learning}

\author{Guoqing Wang$^{1,2}$, Hu Han$^{1,3,}$\thanks{Corresponding author.} , Shiguang Shan$^{1,2,3}$, Xilin Chen$^{1,2}$\\
$^{1}$ Key Laboratory of Intelligent Information Processing of Chinese Academy of Sciences (CAS), \\
Institute of Computing Technology, CAS, Beijing 100190, China,\\
$^{2}$ University of Chinese Academy of Sciences, Beijing 100049, China\\
$^{3}$ Peng Cheng Laboratory, Shenzhen, China\\
{\tt\small guoqing.wang@vipl.ict.ac.cn, \{hanhu, sgshan, xlchen\}@ict.ac.cn}
}
% \affil[1]{ Key Laboratory of Intelligent Information Processing of Chinese Academy of Sciences (CAS), Institute of Computing Technology, CAS, Beijing 100190, China}
% \affil[2]{Peng Cheng Laboratory, Shenzhen, China}
% \affil[3]{University of Chinese Academy of Sciences, Beijing 100049, China}

\maketitle
\newcommand{\RNum}[1]{\uppercase\expandafter{\romannumeral #1\relax}}

%%%%%%%%% ABSTRACT
%%%%%%%%% ABSTRACT
\begin{abstract}
Face presentation attack detection (PAD) has been an urgent problem to be solved in the face recognition systems. Conventional approaches usually assume the testing and training are within the same domain; as a result, they may not generalize well into unseen scenarios because the representations learned for PAD may overfit to the subjects in the training set. In light of this, we propose an efficient disentangled representation learning for cross-domain face PAD. Our approach consists of disentangled representation learning (DR-Net) and multi-domain learning (MD-Net). DR-Net learns a pair of encoders via generative models that can disentangle PAD informative features from subject discriminative features.
The disentangled features from different domains are fed to MD-Net which learns domain-independent features for the final cross-domain face PAD task. Extensive experiments on several public datasets validate the effectiveness of the proposed approach for cross-domain PAD.
\end{abstract}

%%%%%%%%% BODY TEXT
\section{Introduction}
Face recognition (FR) is being widely used in a variety of applications such as smartphone unlock, access control and pay-with-face. The easiness of obtaining a genuine user's face image brings threat to these FR systems, which can be used to presentation attacks (PAs), e.g., photo, video replay, or 3D facial mask. In addition, advances in deep learning have significantly promoted the development of face synthesis technologies like deep facial manipulation attacks \cite{suwajanakorn2017synthesizing}, feasible using Variational AutoEncoders (VAEs) \cite{kingma2013auto}, or Generative Adversarial Network (GAN) \cite{goodfellow2014generative}.

\begin{figure}[t]
\begin{center}
\includegraphics[scale=0.75]{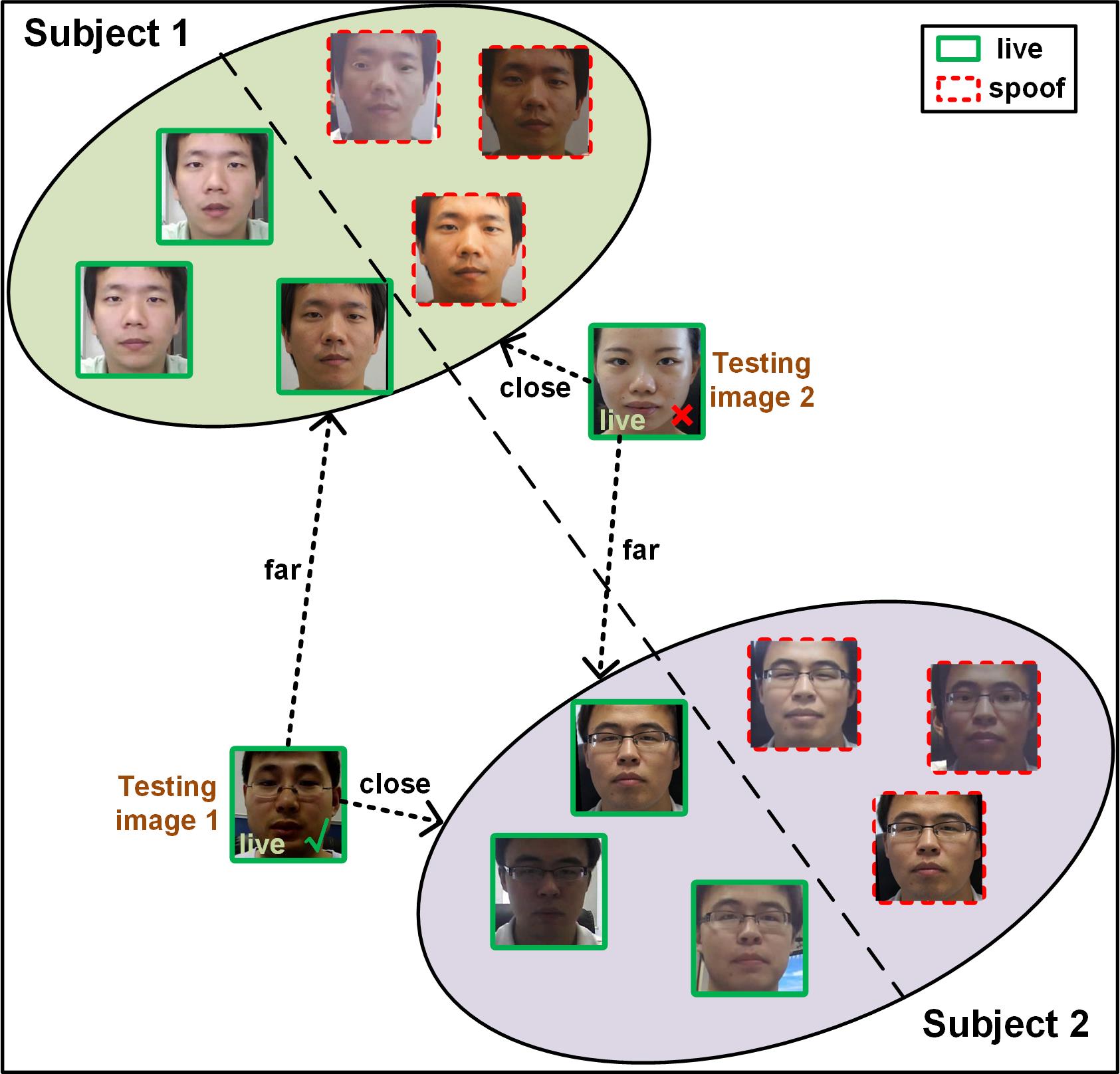}
\end{center}
   \caption{2D visualization of the features learned by ResNet-18 \cite{he2016deep} for live vs. spoof classification on CASIA \cite{zhang2012face}. We observe that the testing face features may not gather together as expected in terms of live and spoof; instead they tend to gather together in terms of individual subjects. This suggests that the features learned for PAD are not disentangled from the features for subject classification very well. We believe this is one of the reasons that leads to poor generalization ability of PAD models under new application scenarios.}
\label{fig1}
\end{figure}
Many approaches have been proposed to tackle various face PAs, which assume that there are inherent disparities between live and spoof faces, such as texture in color space \cite{boulkenafet2015face,boulkenafet2017face}, image distortion \cite{wen2015face}, temporal variation \cite{shao2017deep} or deep semantic features \cite{yang2014learn, patel2016cross}. Although these methods show promising performance under intra-database testing scenarios, their performance dramatically degrades when used in new application scenario (cross-domain PAD). To improve the robustness of PAD under unseen scenarios, some scenario-invariant auxiliary information (such as face depth and heart rhythm) has been exploited for distinguishing between live and spoof faces \cite{atoum2017face, liu2018learning, liu2018remote}. The robustness of these methods relies on the accuracy of the estimated auxiliary information to some extent. With the success of transfer learning in many object recognition tasks \cite{csurka2017domain}, recent studies for PAD seek to utilize transfer learning to achieve end-to-end cross-domain face PAD model or robust features followed by binary classification models \cite{li2018unsupervised, guoqing19ada, shao2019multi}. However, the generalization ability of these methods is still limited when the distributions of data in training and testing domains have big disparities, particularly when the spoof attack types in the testing domain do not appear in the training domain. One important reason is that these methods may not disentangle the domain-independent PAD cues from the domain-dependent cues very well. As shown in Fig. \ref{fig1}, the learned features for face PAD may overfit to subject classification, and thus the live face images (or the spoof face images) of different subjects may not be close to each other in the feature space. In this case, we cannot expect the model learned from a source domain can generalize well to unseen domains. While it could be helpful to improve the cross-domain robustness by collecting a huge face PAD dataset containing various subjects and spoof attack types like the MS-Celeb-1M dataset \cite{guo2016ms} for FR, it can be very expensive and time-consuming in practice.

In this paper, we hope on improving PAD generalization ability without collecting a huge face PAD dataset, and propose a disentangled representation learning approach for cross-domain PAD. In particular, the proposed approach consists of a disentangled representation learning module (DR-Net) and a multi-domain learning module (MD-Net). DR-Net leverages generative models to learn a pair of disentangled feature learning encoders for learning subject classification (denoted as ID-GAN), and PAD (denoted as PAD-GAN) features respectively in each source domain.
The discriminators of PAD-GAN are discriminative not only for distinguishing between the generated and real face images, but also distinguishing between live and spoof face images. Similarly, the discriminators of ID-GAN is discriminative for distinguishing between individual subjects. Thus, with DR-Net, we can expect to obtain features that are either related to PAD or subject classification. On the other hand, we also want to utilize the knowledge about different PA types to enhance the generalization capability of the model. To achieve this goal, we propose a multi-domain feature learning module (MD-Net) to learn domain-independent features from the disentangled features by DR-Net.

The main contributions of this work are as follows: (i) we propose a novel disentangled representation learning for cross-domain PAD, which is able to disentangle features informative for PAD from the features informative for subject classification, allowing more subject-independent PAD feature learning; (ii) we propose an effective multi-domain feature learning to obtain domain-independent PAD features, which enhances the robustness the PAD model; and (iii)
our approach achieves better performance than the state-of-the-art face PAD methods in cross-domain PAD.
% This leads to the model learned from the traininng domain discriminating against the spoof and live faces of the same identity, and the discriminating power of the identity that has not appeared in the test domain.

% The faces with the similar identities from different domains have a small distance in feature space, while the faces with the distinctive identities  has a large distance in feature space.

%-------------------------------------------------------------------------

\section{Related Work}

\textbf{Face presentation attack detection.} Recent face PAD approaches can be generally grouped into conventional approaches, deep learning approaches and domain generalized approaches. Conventional approaches aim to detect attacks based on texture or temporal cues. An early work on face PAD by Li et al. \cite{li2004live} used Fourier spectra analysis to capture the difference between the live and spoof face images. After that, various hand-crafted features such as LBP \cite{maatta2011face}, LPQ \cite{boulkenafet2017face}, SURF \cite{boulkenafet2017face}, HoG \cite{komulainen2013context}, SIFT \cite{patel2016secure} and IDA \cite{wen2015face}, have been utilized with traditional classifiers, e.g., SVM \cite{suykens1999least} and LDA \cite{duda2012pattern}, for live vs. spoof face classification. To reduce the influence by illumination variations, some approaches convert face images from RGB color space to other space such as HSV and YCbCr \cite{boulkenafet2015face,boulkenafet2017face}, and then extract the above features for PAD. Besides, a number of methods also explored temporal cues of the whole face or individual face components for PAD, such as eye blink \cite{sun2007blinking}, mouth movement \cite{kollreider2007real} and head rotation \cite{bao2009liveness}. The conventional approaches are usually computationally efficient and explainable, and work well under intra-database testing scenarios. However, their generalization ability into new application scenarios is still not satisfying \cite{patel2016cross, patel2016secure}.

\begin{figure*}
\begin{center}
%\fbox{\rule{0pt}{2in} \rule{.9\linewidth}{0pt}}
\includegraphics[scale=0.85]{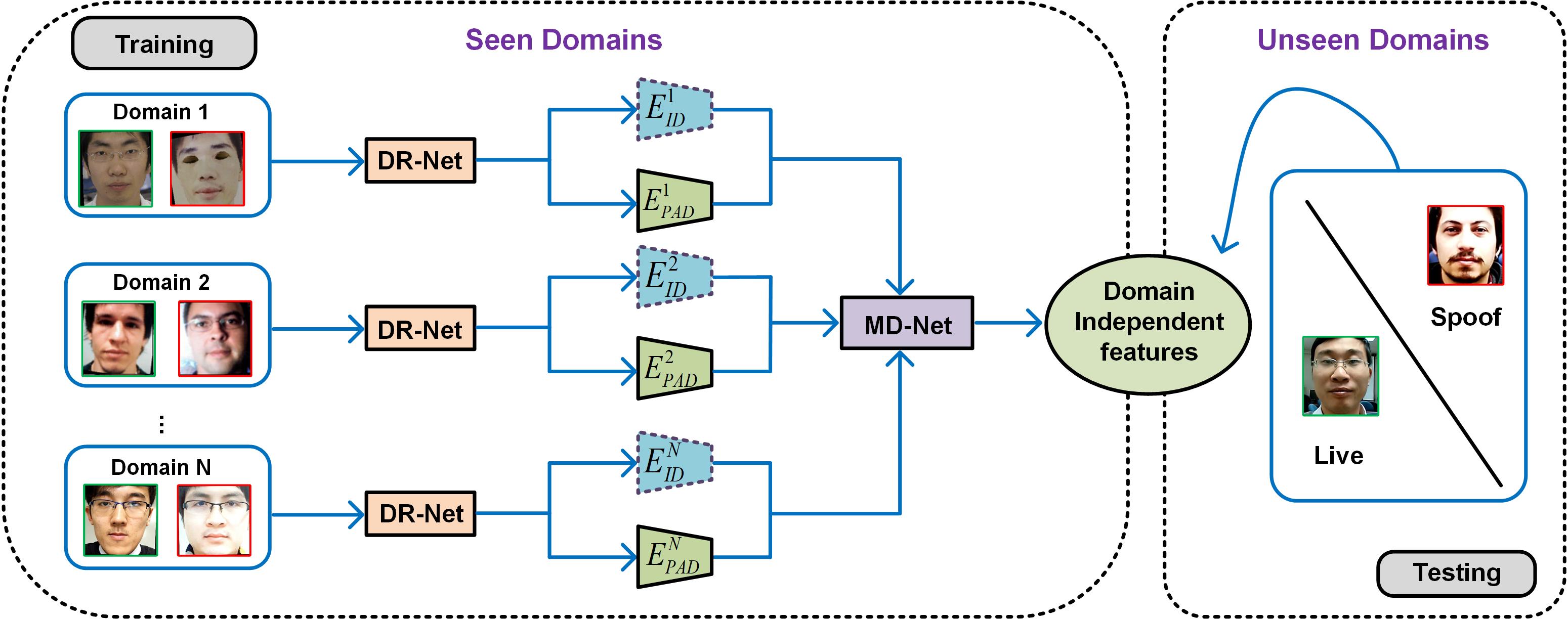}
\end{center}
   \caption{ The overview of our approach for cross-domain PAD. Our approach consists of a disentangled representation learning module (DR-Net) and a multi-domain feature learning module (MD-Net). With the face images from different domains as inputs, DR-Net can learn a pair of encoders for disentangled features for PAD and subject classification respectively. The disentangled features are fed to MD-Net to learn domain-independent representations for robust cross-domain PAD.}
\label{fig2}
\end{figure*}

To overcome the over-fitting issue, researchers are attempting to utilize deep learning models for face PAD in a general-to-specific model transfer scheme \cite{yang2014learn, patel2016cross} because of the great success of deep learning in many other computer vision tasks. In \cite{yang2014learn}, CNNs were fine-tuned with live and spoof face images for face PAD. Liu et al. \cite{liu2018learning} introduced depth map and rPPG signal estimation from RGB face videos as auxiliary supervision to assist in PAD. Jourabloo et al. \cite{jourabloo2018face} inversely decomposed a spoof face into a live face and a noise of spoof, and then utilized the spoof noise for PAD. Zhang et al. \cite{zhang2019dataset} introduced a large-scale multi-modal face anti-spoofing dataset namely CASIA-SURF. Based on CASIA-SURF, several end-to-end approaches \cite{Parkin_2019_CVPR_Workshops, Shen_2019_CVPR_Workshops, wang2019multi} were proposed to exploit the complementary information from RGB, depth and IR, and all reported the promising results. While deep learning approaches show strong feature learning ability, they still do not have pleasing poor generalization ability into new scenarios.

Most recently, deep learning based domain generalization approaches for cross-domain face PAD have been proposed. Li et al. \cite{li2018unsupervised} proposed an unsupervised domain adaptation PAD framework to transform the source domain feature space to the unlabeled target domain feature space by minimizing MMD \cite{gretton2012kernel}. Wang et al. \cite{guoqing19ada} proposed an end-to-end learning approach to improve cross-domain PAD generalization capability by utilizing prior knowledge from source domain via adversarial domain adaptation. Shao et. al \cite{shao2019multi} proposed to learn a generalized feature space via a novel multi-adversarial discriminative deep domain generalization framework under a dual-force triplet-mining constraint. Liu et. al \cite{liu2019deep} defined the detection of unknown spoof attacks as ZSFA and proposed a novel deep tree network to partition the spoof samples into semantic sub-groups in an unsupervised fashion. While the prior domain generalization approaches can improve the model's generalization capability, the representations learned for PAD  may still get over-fitting to specific subjects, shooting environment, camera sensors, etc. In light of this, we propose a disentangled representation learning module and multi-domain feature learning module to obtain more discriminative PAD features. 

\textbf{Disentangled representation learning.} Disentangled representation learning aims to design the appropriate objectives to learn disentangled representations. Recently, more and more methods utilize GAN and VAE to learn disentangled representations because of their success in generated tasks. Yan et. al \cite{yan2016attribute2image} studied a novel problem of attribute-conditioned image generation and proposed a solution with CVAEs. Chen et.al \cite{chen2016infogan} proposed InfoGAN, an information-theoretic extension to GAN that is able to learn disentangled representations in an unsupervised manner.
Tran et. al \cite{tran2017disentangled} proposed DR-GAN for pose-invariant face recognition and face synthesis. Hu et. al \cite{hu2018disentangling} proposed an approach to learn image representations that consist of disentangled factors of variation without exploiting any manual labeling or data domain knowledge. The most related work to the problem we study in this paper is InfoGAN. However, unlike InfoGAN, which maximizing Mutual Information between generated images and input codes, we provide explicit disentanglement and control under the PAD and ID constraints.

\section{Proposed Method}
We aim to build a robust cross-domain face PAD model that can mitigate the impact of the subjects, shooting environment and camera settings from the source domain face images. To achieve this goal, we propose an efficient disentangled representation learning for cross-domain face PAD. As shown in Fig. \ref{fig2}, our approach consists of a disentangled representation learning module (DR-Net) and a multi-domain feature learning module (MD-Net). DR-Net leverages generative models (i.e., PAD-GAN and ID-GAN in Fig. \ref{fig2}) to learn a pair of encoders for learning disentangled features for subject classification and PAD respectively in each source domain. Then, the disentangled features from multiple domains are fed to MD-Net to learn domain-independent features for the final cross-domain face PAD.

\subsection{Disentangled Representation Learning via \\ DR-Net}

\begin{figure}
\begin{center}
%\fbox{\rule{0pt}{2in} \rule{.9\linewidth}{0pt}}
\includegraphics[scale=1.2]{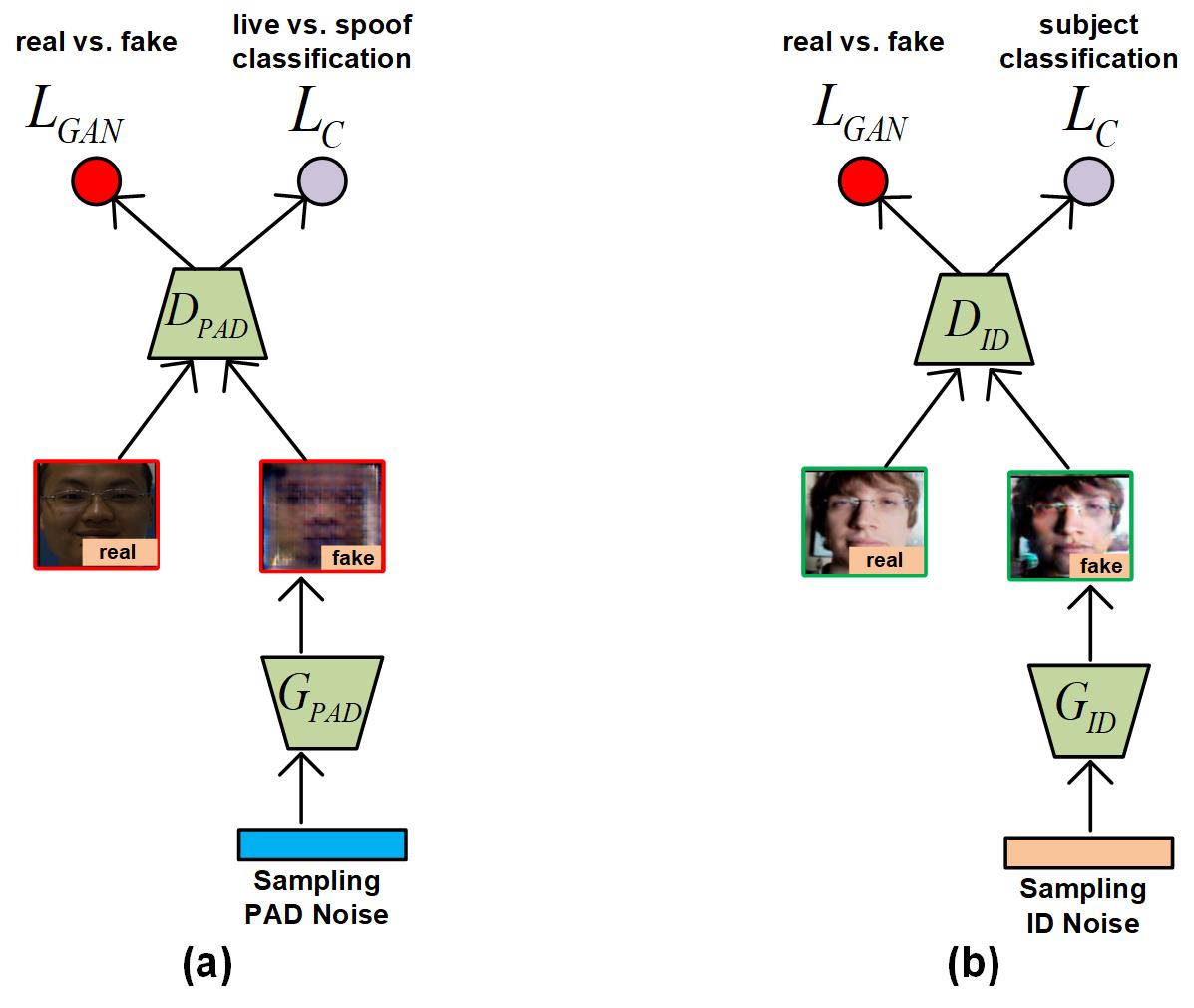}
\end{center}
   \caption{ The diagrams of (a) PAD-GAN and (b) ID-GAN in DR-Net, which can learn disentangled features for PAD and subject classification, respectively. }
\label{fig3}
\end{figure}

As shown in Fig. \ref{fig1}, it is likely to include the subject's identity information in the features learned by CNNs for simple live vs. spoof classification. The main reason is that features related to different tasks can be easily coupled with each other when there are not auxiliary labels for supervising the network learning. Recently, generative models have drawn increasing attentions for obtaining disentangled representation learning when there are no auxiliary labels to use \cite{chen2016infogan, denton2017unsupervised}. The idea is to resemble the real distributions of individual classes by learning a generative model based on the limited data per class. This way, the per class distributions can be modeled in a continuous way instead of defined by a few data points available in the training set. In light of this, we propose a generative DR-Net consisting of PAD-GAN and ID-GAN to learn a pair of disentangled representation encoders for learning subject classification and PAD features respectively in each source domain. As shown in Fig. \ref{fig3}, PAD-GAN and ID-GAN have the same structure, which is a variant of the conventional GAN architecture. So, we only use PAD-GAN as an example to provide the details. ID-GAN is the same as PAD-GAN except for the number of categories in the classification layer. The classification in PAD-GAN handles two classes (live and spoof), while the classification in ID-GAN handles as many classes as the number of subjects in the dataset.
Let $(X,Y)$ denote the face images and its corresponding labels (live vs. spoof) , $z_{c} \in P_{noise}$ denote the input noise vector of PAD-GAN. The generator $G_{PAD}$ of PAD-GAN uses $z_{c}$ to generate face images $X_{fake} = G_{PAD} (z_{c})$. The discriminator $D_{PAD}$ not only needs to predict whether its input face image is real or generated, but also needs to distinguish between live and spoof face images. The loss function of $D_{PAD}$ consists of two parts, i.e., GAN Loss 
\begin{equation}
\label{eq1}
\begin{split}
\mathcal{L}_{GAN} (G_{PAD}, D_{PAD})= \mathbb{E}_{(x, y)\sim (X, Y)}[\log D_{PAD}(x)] \\ 
 + \; \mathbb{E}_{z_{c}\sim P_{noise}}[\log (1 - D_{PAD}(G_{PAD}(z_{c})))], \\
\end{split}
\end{equation}
and classification loss 
\begin{equation}
\label{eq2}
\begin{split}
\mathcal{L}_{C} (G_{PAD}, D_{PAD}& )   = \mathbb{E}_{z_{c}\sim P_{noise}, (x, y)\sim (X, Y)} \\
  [log (D_{PAD} & (G_{PAD}(z_{c}))) ] + [log (D_{PAD}(x))]
\end{split}
\end{equation}
The overall objective of PAD-GAN can be written as 
\begin{equation}
\label{eq3}
\begin{split}
\min_{G_{PAD}} \max_{D_{PAD}} \mathcal{L}_{PAD}  =   \mathcal{L}_{GAN} + \lambda \mathcal{L}_{C}.
\end{split}
\end{equation}
We empirically set $\lambda = 1$ to balance the scales of the GAN loss and classification loss in our experiments. 

With DR-Net, we are able to approximate the real distribution of live vs. spoof face images and the distribution of individual subjects' face images via the encoders in PAD-GAN and ID-GAN, respectively. As a result, the encoders of PAD-GAN and ID-GAN learned in an adversarial way are expected to obtain features that are purely related to PAD and subject classification, respectively. In other words, each encoder can obtain disentangled representation for a specific task. The traditional binary classification methods only utilize some discrete samples to find the boundary in high dimensional feature space. In this way, we may not obtain good disentangled representations. However, we can utilize the generative model to represent the samples from different categories with a continuous distribution. Our experiments in Section \ref{experiment} validate these assumptions.
\begin{figure*}
\begin{center}
%\fbox{\rule{0pt}{2in} \rule{.9\linewidth}{0pt}}
\includegraphics[ scale=0.73]{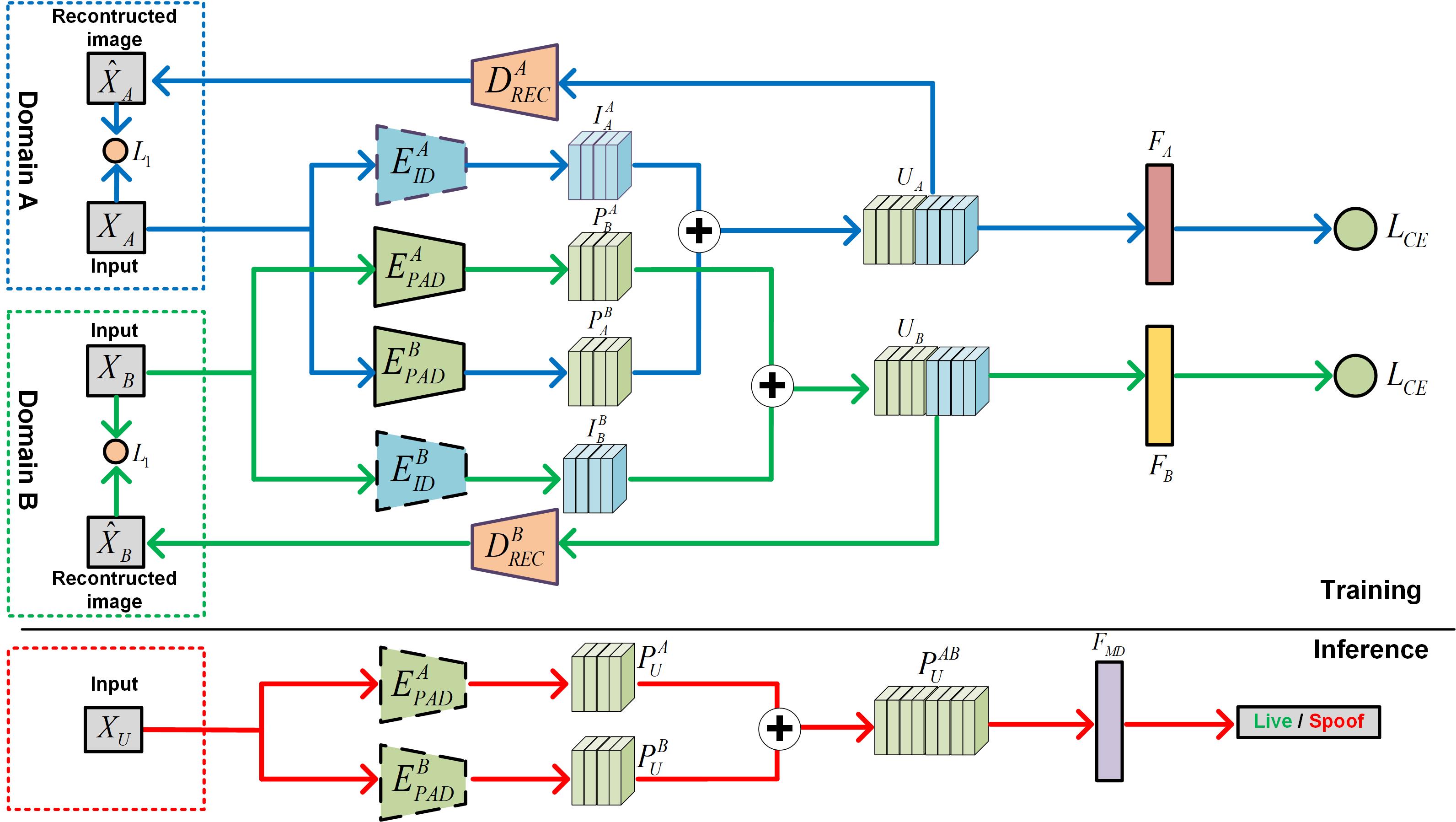}
\end{center}
   \caption{ The overview of MD-Net for domain-independent feature learning. During network training, the two encoders ($E_{ID}^{A}$, $E_{PAD}^{A}$) from domain A and two encoders ($E_{ID}^{B}$, $E_{PAD}^{B}$) from domain B are used to extract disentangled features, i.e., $I_{A}^{A}$, $P_{B}^{A}$, $I_{B}^{B}$, $P_{A}^{B}$. These disentangled features are further enhanced in a cross-verified manner to build concatenated features $U_{A}$ and $U_{B}$, which are used for both subject classification and PAD. The finally learned encoders ($E_{PAD}^{A}$, $E_{PAD}^{B}$) are used to obtain two PAD features per face image, which are concatenated and used to learn the final live vs. spoof classification model to be used during network inference. Dashed lines indicate fixed network parameters.}
\label{fig4}
\end{figure*}
\subsection{Multi-domain Learning via MD-Net}
Given DR-Net, we can obtain a pair of encoders for PAD and subject classification within each domain. Without loss of generality, we assume there are two source domains, and a simple approach in leveraging the knowledge of the two domains to perform cross-domain PAD is to concatenate the features extracted by the PAD encoder of each domain. However, such an approach may not make full use of the disentangled ID features extracted by the ID encoders, which are also helpful for obtaining domain-independent representations for cross-domain PAD. In light of this, we propose a multi-domain feature learning (MD-Net) to fully exploit the disentangled ID and PAD feature representations from different domains. Fig. \ref{fig4} shows the training and inference processes of our MD-Net when learning from two source domains and testing on an unseen domain.

Let $(X_{A}, Y_{A})$ and $(X_{B}, Y_{B})$ denote the face images and the corresponding labels from two domains (A and B), respectively. $E_{ID}^{A}$ and $E_{PAD}^{A}$ denote the ID and PAD encoders learned from $(X_{A}, Y_{A})$. Similarly, $E_{ID}^{B}$ and $E_{PAD}^{B}$ denote the ID and PAD encoders learned from $(X_{B}, Y_{B})$. Then, we can get the PAD features 
\begin{equation}
\label{eq4}
\begin{split}
P_{A}^{B} = \mathbb{E}_{(x, y)\sim (X_{A}, Y_{A})} [E_{PAD}^{B} (x)], \\
P_{B}^{A} = \mathbb{E}_{(x, y)\sim (X_{B}, Y_{B})} [E_{PAD}^{A} (x)],
\end{split}
\end{equation}
where $P_{A}^{B}$ denotes the PAD features extracted by $E_{PAD}^{B}$ for the face images in domain A. Similarly, $P_{B}^{A}$ denotes the PAD features extracted by $E_{PAD}^{A}$ for the face images in domain B. Similarly, we can get the disentangled ID feature representations extracted by $E_{ID}^{A}$ and $E_{ID}^{B}$,
\begin{equation}
\label{eq5}
\begin{split}
I_{A}^{A} = \mathbb{E}_{(x, y)\sim (X_{A}, Y_{A})} [E_{ID}^{A} (x)],
\\
I_{B}^{B} = \mathbb{E}_{(x, y)\sim (X_{B}, Y_{B})} [E_{ID}^{B} (x)].
\end{split}
\end{equation}
We also use the ID features in our MD-Net to enhance the generalization ability of the features for cross-domain PAD. Specifically, as shown in Fig. \ref{fig4}, we leverage the information included in $P_A^B$, $P_B^A$, $I_A^A$, and $I_B^B$ in a cross concatenation manner, and obtain the following features, i.e.,
\begin{equation}
\label{eq6}
\begin{split}
U_{A} = P_{A}^{B} \oplus I_{A}^{A}, 
\\
U_{B} = P_{B}^{A} \oplus I_{B}^{B}. 
\end{split}
\end{equation}
Under the assumptions that the encoders $E_{PAD}$ and $E_{ID}$ can ideally extract PAD and ID features, we expect that the multi-domain feature $U_A$ should retain the identity information from the face image in domain A, and retain the live or spoof information from the face image in domain B. Similarly, the multi-domain feature $U_B$ should retain the identity information from the face image in domain B, and retain the live or spoof information from the face image in domain A. These properties can be used to enhance the learning of $E_{PAD}$. For the property of retaining the live or spoof information, we introduce a live vs. spoof classification branch using classifiers $F_A$ and $F_B$ with cross-entropy loss
\begin{equation}
\label{eq7}
\begin{split}
\mathcal{L}_{CE} &(E_{PAD}^{B}, F_A) = \mathbb{E}_{(x, y)\sim (X_{A}, Y_{A})} \\
[ & - y \log (F_{A}(U_{A})) - (1-y)\log (1 - \log (F_{A}(U_{A})))], \\
\mathcal{L}_{CE} &(E_{PAD}^{A}, F_B ) = \mathbb{E}_{(x, y)\sim (X_{B}, Y_{B})}\\
 [ & - y \log (F_{B}(U_{B})) - (1-y)\log (1 - \log (F_{B}(U_{B}))))].
\end{split}
\end{equation}
For the  property of retaining identity information, we introduce an image reconstruction branch using decoders $D_{REC}^{A}$ and $D_{REC}^{B}$ with $L_{1}$ loss
\begin{equation}
\label{eq8}
\begin{split}
\mathcal{L}_{1} (E_{PAD}^{B}, D_{REC}^{A}) = {\mathbb{E}_{x_{A}\sim X_{A}}}[\left \| x_{A} -D_{REC}^{A} (U_{A}) \right \|_1], \\
\mathcal{L}_{1} (E_{PAD}^{A}, D_{REC}^{B}) = {\mathbb{E}_{x_{B}\sim X_{B}}}[\left \| x_{B} -D_{REC}^{B} (U_{B}) \right \|_1].
\end{split}
\end{equation}
After the cross-verified learning of MD-Net, the encoders $E_{PAD}$ from DR-Net can be expected to have the ability to generate more general features that are preferred for cross-domain PAD. All the above optimizations in MD-Net are performed in a commonly used multi-task learning manner \cite{han2019pami}.

Finally, for all the face images from domains A and B, we use the learned PAD encoders $E_{PAD}^{A}$ and $E_{PAD}^{B}$ to extract two PAD features per face image, which are then concatenated into one feature and used for learning a live vs. spoof classifier $F_{MD}$. Similarly, during inference, given a face image from an unseen domain $X_U$, we use $E_{PAD}^{A}$ and $E_{PAD}^{B}$ to extract two PAD features, i.e., $P_U^A$ and $P_U^B$, and use their concatenation to perform live vs. spoof classification using $F_{MD}$.

\section{Experimental Results}

%-------------------------------------------------------------------------
\subsection{Databases and Protocols}
\begin{table}
\newcommand{\tabincell}[2]{\begin{tabular}{@{}#1@{}}#2\end{tabular}}
\footnotesize
\begin{center}
\begin{tabular}{l|c|c|c|c}
\hline
\hline
\multirow{2}{*}{Dataset} &
\multirow{2}{*}{\tabincell{c}{PA \\ types} } &
\multicolumn{2}{c|}{No. of Subj.} &
\multirow{2}{*}{\tabincell{c}{Display \\ devices}} 
\\
\cline{3-4}
 & & train & test & \\
\hline 
\tabincell{c} {CASIA \cite{zhang2012face}\\(abbr.: C)}  & \tabincell{c}{ Printed photo\\Cut photo\\Replayed video }  & 20 & 30 & iPad \\
\hline
\tabincell{c} {Idiap \cite{chingovska2012effectiveness} \\(abbr.: I)} & \tabincell{c}{ Printed photo\\Display photo\\Replayed video }  & 30 & 20 & \tabincell{c}{iPhone 3GS \\ ipad}    \\
\hline
\tabincell{c}{MSU \cite{wen2015face} \\(abbr.:M)} & \tabincell{c}{ Printed photo\\Replayed video }  & 18 & 17 & \tabincell{c}{iPad Air \\ iphone 5S}  \\
\hline
\tabincell{c}{OULU \cite{OULU_NPU_2017} \\ (abbr.:O)} & \tabincell{c}{ Printed photo\\Display photo\\Replayed video }  & 35 & 20 & \tabincell{c}{Dell 1905FP \\ Macbook Retina}  \\ 
\hline
\hline
\end{tabular}
\end{center}
\caption{A summary of the PAD databases used in our experiments.}
\label{dataset}
\end{table}
We provide evaluations on four widely used databases for cross-domain PAD including Idiap REPLAY-ATTACK \cite{chingovska2012effectiveness} (Idiap), CASIA Face AntiSpoofing \cite{zhang2012face} (CASIA), MSU-MFSD \cite{wen2015face} (MSU) and OULU-NPU \cite{OULU_NPU_2017} (OULU). Table \ref{dataset} provides a summary of the four datasets in terms of the PA types, display devices, and the number of subjects in the training and testing sets. 

We regard one dataset as one domain in our experiment. Given the four datasets (Idiap, CASIA, MSU, OULU), we can define three types of protocols to validate the effectiveness of our approach. In \textbf{protocol \RNum{1}}, we train the model using all the images from three datasets, and test the model on the fourth dataset (denoted as [A, B, C]$\to$D). Then, we have four tests in total for protocol \RNum{1}: [O, C, I]$\to$M, [O, M, I]$\to$C, [O, C, M]$\to$I, [I, C, M]$\to$O. In \textbf{protocol \RNum{2}}, we train the model using all the images from two datasets, and test the model on a different dataset (denoted as [A, B]$\to$C). So, there are 12 tests in total for protocol \RNum{2}, e.g., [M, I]$\to$C, [M, I]$\to$O, [M, C]$\to$I and [M, C]$\to$O, etc. In protocol \textbf{\RNum{3}}, we train the model using only the images from one dataset, and test the model on a different dataset (denoted as A$\to$B). Then, there are also 12 tests in total for protocol \RNum{3}, e.g., I$\to$C, I$\to$M, C$\to$I and C$\to$M, etc.

% follow the same testing protocol as that in \cite{shao2019multi}, i.e., train the model using all the images from datasets A, B and C, and test the model on a different dataset D (denoted
% as A \& B \& C $\to$ D). So, we have four tests in total: O \& C \& I $\to$ M, O \& M \& I $\to$ C, O \& C \& M $\to$ I, I \& C \& M $\to$ O, in which C, M, I and O denote CASIA, MSU, Idiap and OULU, respectively.

% 这里把从3个数据库到1个数据库的测试定义为：Prococol I, 然后累出有哪几种从3到1测组合；应该是4种吧。
% 把从2个数据库到1个数据库的测试定义为： Protocol II, 类似的给出从2个到1个组合。

%-------------------------------------------------------------------------

\subsection{Baselines and Evaluation Metric}

We compare our approach with several state-of-the-art methods that can work for cross-domain PAD, e.g., Auxiliary-supervision (Aux) \cite{liu2018learning} using CNN and RNN \cite{hochreiter1997long} to jointly estimate the face depth and rPPG signal from face video, MMD-AAE \cite{li2018domain} learning a feature representation by jointly optimization a multi-domain autoencoder regularized by the MMD \cite{gretton2012kernel} distance, a discriminator and a classifier in an adversarial training manner, and MADDG \cite{shao2019multi} learning a generalized feature space by training one feature generator to compete with multiple domain discriminators simultaneously.

We also use a number of conventional PAD methods as baselines, such as Multi-Scale LBP (MS\_LBP) \cite{maatta2011face}, Binary CNN (CNN)\cite{yang2014learn}, Image Distortion Analysis (IDA) \cite{wen2015face}, Color Texture (CT) \cite{boulkenafet2016face} and LBPTOP \cite{de2014face}.

We follow the state-of-the-art methods for cross-domain face PAD \cite{li2018unsupervised, li2018learning, guoqing19ada} and report Half Total Error Rate (HTER) \cite{anjos2011counter} (the average of false acceptance rate and false rejection rate) for cross-domain testing. We also report Area Under Curve (AUC) in order to compare with baseline methods such as \cite{shao2019multi}.

\subsection{Implementation Details}

\textbf{Network Structure.} The PAD-GAN and ID-GAN in DR-Net share the same structure. To be specific, the generators have one FC layer and seven transposed convolution layers to generate fake images of $256 \times 256 \times 3$ from the 512-D noise vectors. There are four residual blocks in the discriminators and each block has four convolution layers, which have the same settings as the convolution part of the ResNet-18 \cite{he2016deep}. The discriminators learned from DR-Net are then used as the disentangled feature encoders to extract the features used by MD-Net. The decoders in MD-Net also have seven transposed convolution layers to generate the reconstructed images of $256 \times 256 \times 3$ from the 2048-D multi-domain features. We use a kernel size of 3 for all transposed convolution layers and use ReLU \cite{nair2010rectified} for activation.

\textbf{Training Details.} We use an open source SeetaFace $\footnote{https://github.com/seetaface/SeetaFaceEngine}$ algorithm to do face detection and landmark localization. All the detected faces are then normalized to $256\times256$ based on five facial keypoints (two eye centers, nose, and two mouth corners). We also use an open source imgaug$\footnote{https://github.com/aleju/imgaug}$ library to perform data augmentation, i.e., random flipping, rotation, resizing, cropping and color distortion. The proposed approach is trained end-to-end and divided into two stages to make the training process stable. At stage-1, we use the face images from different domains to train DR-Net with the joint of GAN loss and classification loss. We choose Adam with the fixed learning rate of $2e^{-3}$, and a batch size of 128 to train the generators and discriminators. At stage-2, we use the discriminators learned from DR-Net as the disentangled feature encoders to learn MD-Net. We also choose Adam as the optimizers of the encoders and decoders with initial learning rates of $1e^{-5}$ and $1e^{-4}$. We train 300 and
100 epochs for the two stages, respectively.
%-------------------------------------------------------------------------

\subsection{Results} \label{experiment} 

% \subsubsection{Cross-domain Testing}

\begin{table*}
\newcommand{\tabincell}[2]{\begin{tabular}{@{}#1@{}}#2\end{tabular}}
\begin{center}
\footnotesize
\begin{tabular}{l|c|c|c|c|c|c|c|c}
\hline
\multirow{2}{*}{Method} &
\multicolumn{2}{c|}{[O, C, I]$\to$M } &
\multicolumn{2}{c|}{[O, M, I]$\to$C} &
\multicolumn{2}{c|}{[O, C, M]$\to$I} &
\multicolumn{2}{c}{[I, C, M]$\to$O}
\\
\cline{2-9}
 & HTER(\%) & AUC(\%) & HTER(\%) & AUC(\%) & HTER(\%) & AUC(\%) & HTER(\%) & AUC(\%) \\
\hline
MS\_LBP \cite{maatta2011face}  & 29.76 & 78.50 & 54.28 & 44.98 & 50.30 & 51.64 & 50.29 & 49.31 \\
CNN \cite{yang2014learn}  & 29.25 & 82.87 & 34.88 & 71.95 & 34.47 & 65.88 & 29.61 & 77.54 \\
IDA \cite{wen2015face} & 66.67 & 27.86 & 55.17 & 39.05 & 28.35 & 78.25 & 54.20 & 44.59 \\
CT \cite{boulkenafet2016face} & 28.09 & 78.47 & 30.58 & 76.89 & 40.40 & 62.78 & 63.59 & 32.71 \\
LBPTOP \cite{de2014face} & 36.90 & 70.80 & 42.60 & 61.05 & 49.45 & 49.54 & 53.15 & 44.09 \\
Aux(Depth only) \cite{liu2018learning} & 22.72 & 85.88 & 33.52 & 73.15 & 29.14 & 71.69 & 30.17 & 66.61 \\
Aux(All) \cite{liu2018learning}  & - & - & 28.4 & - & 27.6 & - & - & - \\
MMD-AAE \cite{li2018domain}  & 27.08 & 83.19 & 44.59 & 58.29 & 31.58 & 75.18 & 40.98 & 63.08 \\
MADDG \cite{shao2019multi} & 17.69 & 88.06 & 24.5 & 84.51 & 22.19 & 84.99 & 27.98 & 80.02 \\
\hline
Proposed approach  & {\bfseries 17.02} & {\bfseries 90.10} & {\bfseries 19.68} & {\bfseries 87.43} & {\bfseries 20.87} & {\bfseries 86.72} & {\bfseries 25.02} & {\bfseries 81.47} \\
\hline
\end{tabular}
\end{center}
\caption{Cross-domain PAD performance of the proposed approach and the baseline methods under \textbf{protocol \RNum{1}}.}
\label{baseline}
\end{table*}

\begin{table}
\newcommand{\tabincell}[2]{\begin{tabular}{@{}#1@{}}#2\end{tabular}}
\begin{center}
\footnotesize
\begin{tabular}{l|c|c|c|c}
\hline
\multirow{2}{*}{Method} &
\multicolumn{2}{c|}{[M, I]$\to$C} &
\multicolumn{2}{c}{[M, I]$\to$O}
\\
\cline{2-5}
 & HTER(\%) & AUC(\%) & HTER(\%) & AUC(\%)  \\
\hline
MS\_LBP \cite{maatta2011face} & 51.16 & 52.09 & 43.63 & 58.07 \\
IDA \cite{wen2015face} & 45.16 & 58.80 & 54.52 & 42.17 \\
CT \cite{boulkenafet2016face} & 55.17 & 46.89 & 53.31 & 45.16 \\
LBPTOP \cite{de2014face}  & 45.27 & 54.88 & 47.26 & 50.21 \\
MADDG \cite{shao2019multi} & 41.02 & 64.33 & 39.35 & 65.10  \\
\hline
Proposed approach & {\bfseries 31.67} & {\bfseries 75.23} & {\bfseries 34.02} & {\bfseries 72.65}  \\
\hline
\end{tabular}
\end{center}
\caption{Cross-domain PAD performance of the proposed approach and the baseline methods under \textbf{protocol \RNum{2}}.}
\label{limited baseline}
\end{table}
\begin{table*}
\begin{center}
\footnotesize
\begin{tabular}{l|c|c|c|c|c|c|c|c|c|c|c|c}
\hline
Method & C $\to$ I & C $\to$ M & I $\to$ C & I $\to$ M & M $\to$ C & M $\to$ I & O $\to$ I & O $\to$ M & O $\to$ C & I $\to$ O & M $\to$ O & C $\to$ O  \\
\hline\hline
CNN \cite{yang2014learn} & 45.8 & 25.6 & 44.4 & 48.6 & 50.1 & 49.9 & 47.4 & 30.2 & 41.2 & 45.4 & 31.4 & 36.4 \\
SA$^\S$ \cite{li2018unsupervised} & 39.2 & 14.3 & 26.3 & 33.2 & 10.1 & 33.3 & - & -
& - & - & - & -\\
KSA$^\S$ \cite{li2018unsupervised} & 39.3 & 15.1 & {\bfseries 12.3} & 33.3 & {\bfseries 9.1} & 34.9 & - & -
& - & - & - & - \\
ADA \cite{guoqing19ada} & {\bfseries 17.5} & {\bfseries 9.3} & 41.6 & 30.5 & 17.7 & {\bfseries 5.1} & {\bfseries 26.8} & 31.5 & {\bfseries 19.8} & 39.6 & {\bfseries 31.2} & 29.1 
\\
\hline
Proposed (PAD-GAN) & 26.1 & 20.2 & 39.2 & {\bfseries 23.2} & 34.3 & 8.7 & 27.6 & {\bfseries 22.0} & 21.8 & {\bfseries 33.6} & 31.7 & {\bfseries 24.7}    \\
\hline
\end{tabular}
\end{center}
\caption{HTER in (\%) of cross-domain PAD by the proposed approach and baseline domain adaptation approaches under \textbf{protocol \RNum{3}}.}
\label{DA-baseline}
\end{table*}

% Aux \cite{liu2018learning} & 27.6 & - & 28.4 & - & - & - & - & - & - & - & - & -
% \\
% De-spoof \cite{jourabloo2018face} & 28.5 & - & 41.1 & - & - & - & - & - & - & - & - & -
% \\
% \begin{table*}
% \begin{center}
% \footnotesize
% \begin{tabular}{l|c|c|c|c|c|c|c|c|c|c|c|c|c}
% \hline
% Method & C $\to$ I & C $\to$ M & I $\to$ C & I $\to$ M & M $\to$ C & M $\to$ I & O $\to$ I & O $\to$ M & O $\to$ C & I $\to$ O & M $\to$ O & C $\to$ O & Avg \\
% \hline\hline
% CNN \cite{yang2014learn} & 45.8 & 25.6 & 44.4 & 68.6 & 50.1 & 49.9 & 47.4 \\
% SA$^\S$ \cite{li2018unsupervised} & 39.2 & 14.3 & 26.3 & 33.2 & 10.1 & 33.3 & 26.1\\
% KSA$^\S$ \cite{li2018unsupervised} & 39.3 & 15.1 & 12.3 & 33.3 & {\bfseries 9.1} & 34.9 & 24.0\\
% ADA \cite{guoqing19ada} & {\bfseries 17.5} & {\bfseries 9.3} & 41.6 & 30.5 & 17.7 & {\bfseries 5.1} & {\bfseries 20.3} \\
% \hline
% Proposed (PAD-GAN) & 26.1 & 20.2 & 39.2 & {\bfseries 23.2} & 34.3 & 8.7 & 25.3  \\
% \hline
% \end{tabular}
% \end{center}
% \caption{HTER of cross-domain PAD in (\%) of the proposed approach and baseline domain adaptation approaches.}
% \label{DA-baseline}
% \end{table*}
\begin{table*}
\footnotesize
\begin{center}
\begin{tabular}{l|c|c|c|c|c|c|c|c}
\hline
\multirow{2}{*}{Method} &
\multicolumn{2}{c|}{[O, C, I]$\to$M} &
\multicolumn{2}{c|}{[O, M, I]$\to$C} &
\multicolumn{2}{c|}{[O, C, M]$\to$I} &
\multicolumn{2}{c}{[I, C, M]$\to$O}
\\
\cline{2-9}
 & HTER(\%) & AUC(\%) & HTER(\%) & AUC(\%) & HTER(\%) & AUC(\%) & HTER(\%) & AUC(\%) \\
\hline
Proposed w/o DR\&MD & 30.25 & 78.43 & 33.97 & 72.32 & 35.21 & 65.88 & 28.74 & 74.27\\
Proposed w/o DR & 26.43 & 84.52 & 31.43 & 77.62 & 29.56 & 72.38 & 27.23 & 74.67\\
Proposed w/o MD & 18.07 & 87.26 & 24.85 & 85.34 & 22.44 & 85.96 & 26.78 & 76.82\\
\hline
Proposed w/o $L_{CE}$ & 18.07 & 87.84 & 24.27 & 85.64 & 22.21 & 86.23 & 26.77 & 76.23\\
Proposed w/o $L_{REC}$ & 18.14 & 88.41 & 21.03 & 86.30 & 21.47 & 86.26 & 26.21 & 78.45\\
\hline
Proposed (full method) & {\bfseries 17.02} & {\bfseries 90.10} & {\bfseries 19.68} & {\bfseries 87.43} & {\bfseries 20.87} & {\bfseries 86.72} & {\bfseries 25.02} & {\bfseries 81.47} \\
\hline
\end{tabular}
\end{center}
\caption{Ablation study of the proposed approach in terms of DR-Net, MD-Net, $L_{CE}$ and $L_{REC}$ losses.}
\label{ablation} 
\end{table*}
\textbf{Protocol \RNum{1}.} The cross-domain PAD performance by the proposed approach and the state-of-the-art methods under protocol \RNum{1} are shown in Table \ref{baseline}. We can observe that conventional PAD methods usually perform worse than the domain generalized methods like Auxiliary-supervision \cite{liu2018learning}, MMD-AAE \cite{li2018domain}, MADDG \cite{shao2019multi}. The possible reason why conventional PAD methods do not work well is that the distributions of data in training and testing domains have big disparities, particularly when the spoof attack types, illumination or display devices in the testing domain do not appear in the training domain. Auxiliary supervision like using rPPG signal \cite{niu2018icpr} in a multi-task way \cite{wang2017fg} was introduced in \cite{liu2018learning}, and reported better performance than simply using CNN for binary classification \cite{yang2014learn}. This is understandable if we look at the feature visualization in Fig. \ref{fig1}. MADDG \cite{shao2019multi} and our approach perform better than Auxiliary-supervision \cite{liu2018learning} which suggests that deep domain generalization is useful for cross-domain PAD. However, our approach performs better than MADDG \cite{shao2019multi} and MMD-AAE \cite{li2018domain} because of multi-domain disentangled representation learning.

\textbf{Protocol \RNum{2}.} The part of cross-domain PAD performance of the proposed approach and baseline methods under protocol \RNum{2} are shown in Table \ref{limited baseline}. We notice that our approach also shows the promising results compared to the conventional approaches and MADDG \cite{shao2019multi}. This suggests that our approach is more able to exploit domain-independent features for cross-domain PAD than MADDG \cite{shao2019multi}. The possible reason is that our approach utilizes the generative model to disentangle the identity information from the face images. This way may be better than the dual-force triplet-mining constraint proposed in \cite{shao2019multi} to eliminate identity information. By comparing the cases when CASIA and OULU are the testing sets in Table \ref{baseline} and Table \ref{limited baseline}, we get the conclusion that more source domains are available, and thus the advantage of domain generalization can be better exploited by our approach. 

\textbf{Protocol \RNum{3}.} 
The cross-domain PAD performance by the proposed approach and baseline  domain adaptation methods under protocol \RNum{3} are shown in Table \ref{DA-baseline}. We can observe that PAD-GAN has a comparable performance with the domain adaptation methods and a better performance than simply using CNN for binary classification \cite{yang2014learn}. This further demonstrates that disentangled representation learning is more potential than pure representation learning when the images are coupled with diverse information. However, the domain adaptation methods perform better than PAD-GAN in some cases, especially using CASIA as a training or testing set. The possible reason is that CASIA has cut photos which are not appeared in other datasets. The domain adaptation methods can alleviate this issue by learning a common feature space shared by both the source and target domains.

\begin{figure}
\begin{center}
%\fbox{\rule{0pt}{2in} \rule{.9\linewidth}{0pt}}
\includegraphics[scale=1.2]{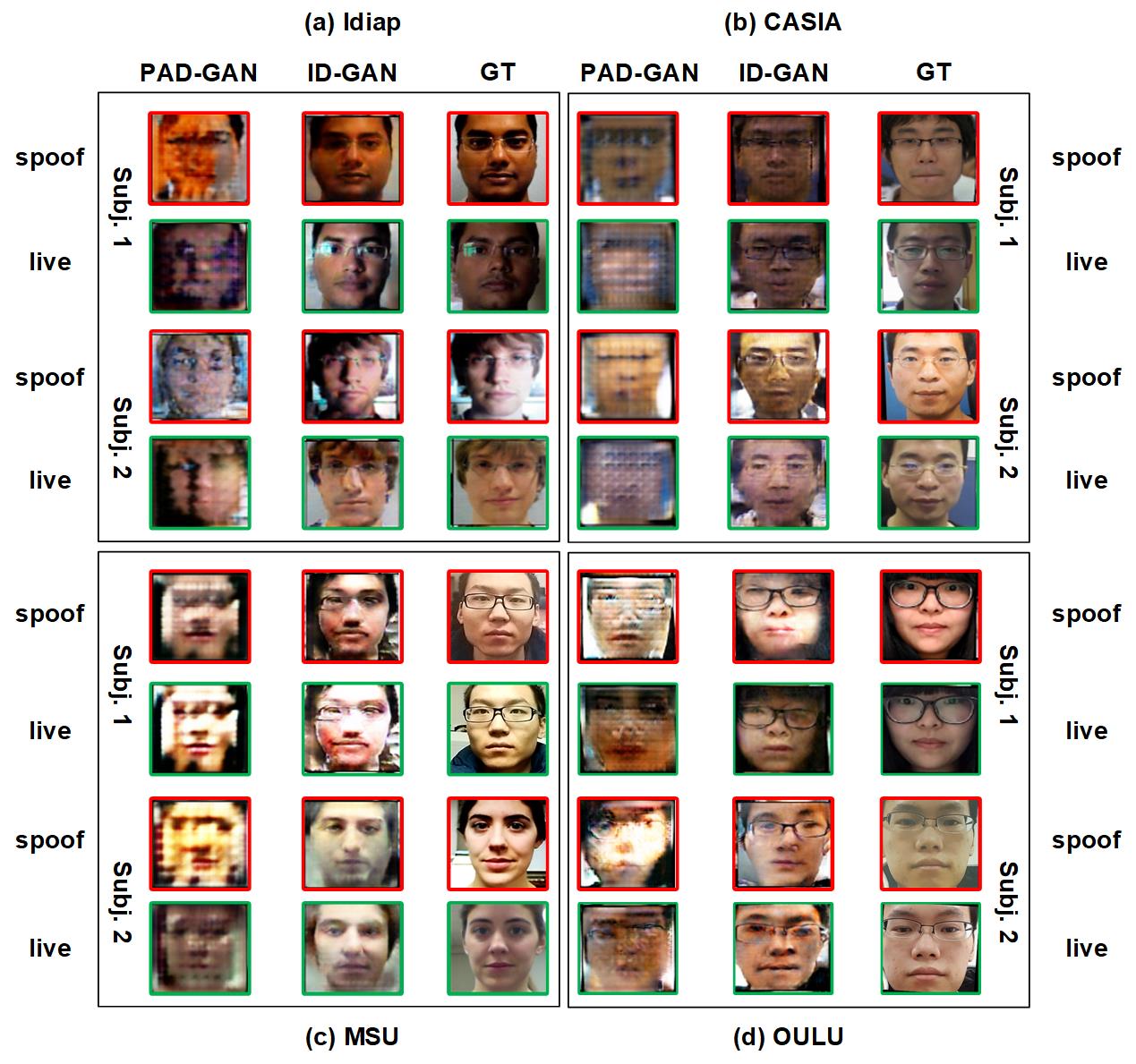}
\end{center}
   \caption{Examples of generated results by the proposed DR-Net on (a) Idiap, (b) CASIA, (c) MSU and (d) OULU. The first two columns list the results generated by PAD-GAN and ID-GAN. The third column lists the results of the ground-truth.}
\label{fig5}
\end{figure}

\begin{figure}
\begin{center}
%\fbox{\rule{0pt}{2in} \rule{.9\linewidth}{0pt}}
\includegraphics[scale=0.7]{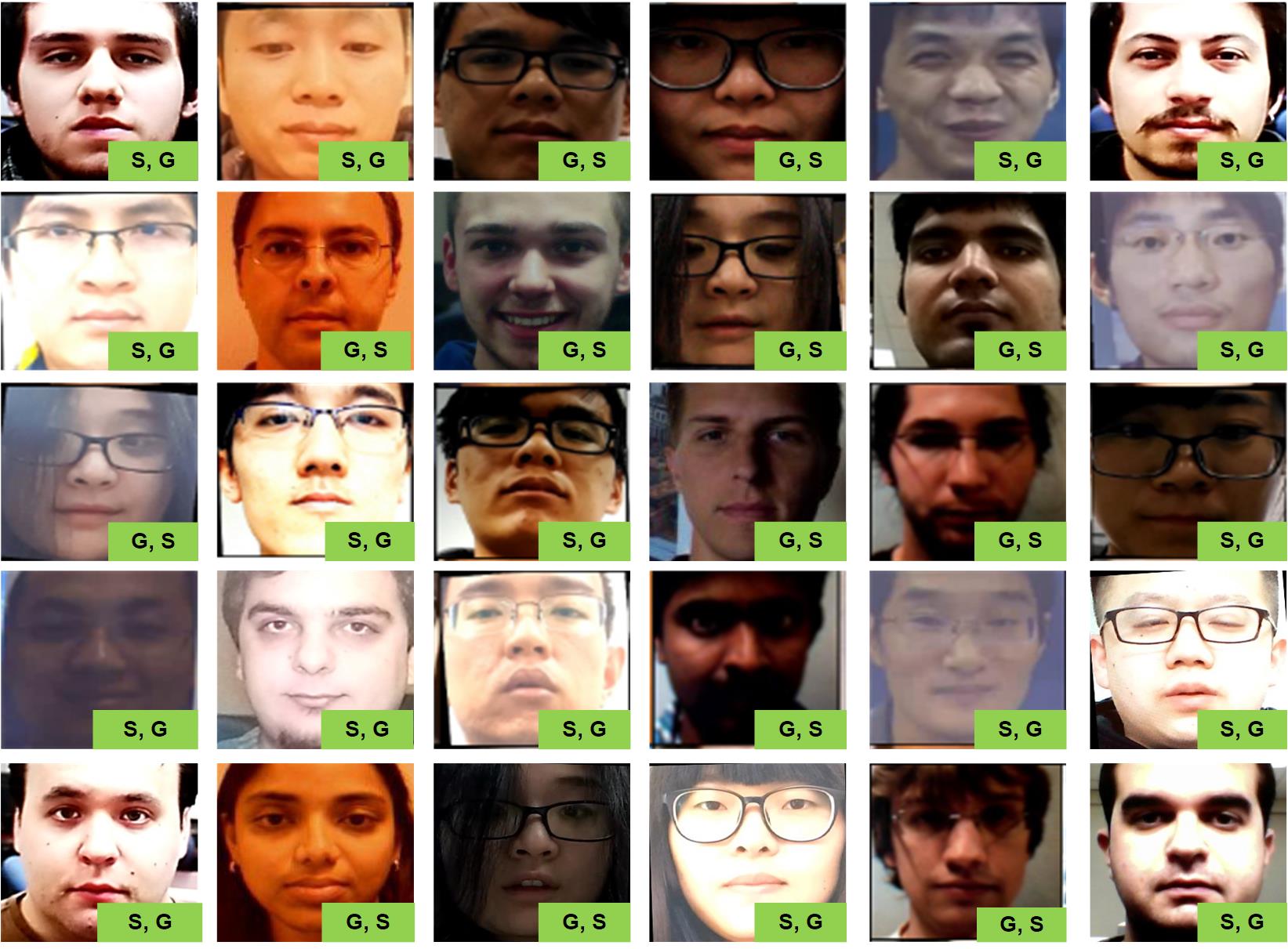}
\end{center}
   \caption{Examples of incorrect PAD results by the proposed approach on Idiap, CASIA, MSU and OULU. The label ``S, G'' (or ``G, S'') denotes a spoof (or live) face image is incorrectly classified as live (or spoof) face image.}
\label{fig6}
\end{figure}

\textbf{Visualization and Analysis.} 
We also visualize the generated fake face images by PAD-GAN and ID-GAN in DR-Net. As shown in Fig. \ref{fig5}, we notice that it is hard to tell the identity of the fake face images generated by PAD-GAN (Fig. \ref{fig3} (a)), and we can only see a rough face shape. By contrast, the fake face images generated by ID-GAN can better keep the identity information than PAD-GAN. This demonstrates the usefulness of DR-Net, which can disentangle the ID and PAD features well from the face images. This again validates that PAD-GAN and ID-GAN in our DR-Net are effective in learning disentangled features for live vs. spoof classification and subject classification, respectively. Such a characteristic is very important for improving cross-domain PAD robustness. Fig. \ref{fig6} shows some examples of incorrect PAD results by the proposed approach on four datasets. We notice that most errors are caused by the challenging appearance variances due to over-saturated or under-exposure illumination, color distortions, or image blurriness that diminishes the differences between live and spoof face images. 


\subsection{Ablation Study}
% \textbf{Effectiveness of DR-Net:} In order to illustrate the effectiveness of the disentangled representation learning module (DR-Net), we conduct the experiments with and without using PAD-GAN to train the PAD encoders. When the PAD-GAN is not used, the  

\textbf{Effectiveness of DR-Net and MD-Net.}
We study the influences of DR-Net and MD-Net by gradually dropping them from the proposed approach, and denote the corresponding models as `Proposed w/o DR', `Proposed w/o MD' and `Proposed w/o DR \& MD'. The results for cross-domain PAD are given in Table \ref{ablation}. We can consider ‘CNN’ in Table \ref{baseline} is the method without disentangling identity information, while ‘w/o MD’ and ‘full method’ in Table \ref{ablation} are methods with partial and complete identity information disentanglement, respectively. We can see identity information disentanglement is useful for improving generalization ability. Furthermore, we can see dropping either component can lead to performance drop. This suggests that both components are useful in the proposed face PAD approach. 

\textbf{Effectiveness of $L_{REC}$ and $L_{CE}$ in MD-Net.} We also conduct the experiments by gradually removing $L_{CE}$ and $L_{REC}$ in MD-Net to validate their usefulness. We denote the corresponding model as `Proposed w/o $L_{CE}$' and `Proposed w/o $L_{REC}$'. The results for cross-domain PAD are given in Table \ref{ablation}. We can see that removing either loss can lead to performance drop. However, $L_{CE}$ has a greater influence to the cross-domain PAD performance than $L_{REC}$.

\section{Conclusions}
Cross-domain face presentation attack detection remains a challenging problem due to diverse presentation attack types and environmental factors. We propose an effective disentangled representation learning for cross-domain presentation attack detection, which consists of disentangled representation learning (DR-Net) and multi-domain feature learning (MD-Net). DR-Net leverages generative model to obtain better modeling of the live and spoof class distributions, allowing effective learning of disentangled features for PAD. MD-Net further learns domain-independent feature representation from the disentangled features for the final cross-domain face PAD task. The proposed approach outperforms the state-of-the-art methods on several public datasets.  

\section{Acknowledgement}
This research was supported in part by the National Key R\&D Program of China (grant 2018AAA0102501), Natural Science Foundation of China (grants 61672496 and 61702481), and Youth Innovation Promotion Association CAS (grant 2018135).

{\small
\bibliographystyle{ieee_fullname}
\bibliography{egpaper_final}
}

\end{document}